\documentclass[conference]{IEEEtran}
\IEEEoverridecommandlockouts
\usepackage{cite}
\usepackage{amsmath,amssymb,amsfonts}
\usepackage{algorithmic}
\usepackage{algorithm}
\usepackage{graphicx}
\usepackage{textcomp}
\usepackage{xcolor}
\usepackage{url}
\usepackage{hyperref}

\def\BibTeX{{\rm B\kern-.05em{\sc i\kern-.025em b}\kern-.08em
    T\kern-.1667em\lower.7ex\hbox{E}\kern-.125emX}}

\hypersetup{
    hidelinks,
    pdftitle={SNR-Gated LSTM-Conditioned Diffusion Model for MIMO Channel Estimation},
    pdfauthor={Jixing Zhou and Xinming Huang}
}

\begin{document}

\title{SNR-Gated LSTM-Conditioned Diffusion Model for MIMO Channel Estimation\\
\thanks{This work was partially supported by The MathWorks Inc. The implementation is available at \url{https://github.com/JixingZhou2/GSL-Diff}
}}

\author{\IEEEauthorblockN{Jixing Zhou}
\IEEEauthorblockA{
%\textit{ Dept. of Electrical and Computer Engineering} \\
\textit{Worcester Polytechnic Institute}\\
Worcester, MA, USA \\
jzhou1@wpi.edu}
\and
\IEEEauthorblockN{Xinming Huang}
\IEEEauthorblockA{
%\textit{ Dept. of Electrical and Computer Engineering} \\
\textit{Worcester Polytechnic Institute}\\
Worcester, MA, USA \\
xhuang@wpi.edu}
% \and
% \IEEEauthorblockN{Amit Kansal, Ying-On Yan}
% \IEEEauthorblockA{
% %\textit{dept. name of organization (of Aff.)} \\
% \textit{The MathWorks Inc.}\\
% Natick, MA, USA \\
% akansal@mathworks.com}
}

\maketitle

\begin{abstract}
Accurate and low latency channel estimation is critical for modern MIMO systems, particularly under mobility, where channels exhibit structured sparsity and strong temporal correlation. This paper proposes a time-series conditioned diffusion framework for channel estimation that performs denoising in the angular domain. Starting from least squares (LS) observations, we train a diffusion denoiser whose conditioning information is encoded by a long short-term memory (LSTM) network over a short observation sequence, enabling the model to exploit temporal dynamics beyond per-snapshot estimation. To robustly balance observation fidelity and learned generative priors across a wide signal-to-noise ratio (SNR) range, we introduce a learnable SNR-gated late-fusion shortcut that injects the network input into the final decoding stage through a sigmoid gate with trainable center and scale. To reduce inference latency, we adopt deterministic denoising diffusion implicit model (DDIM) style reverse updates with SNR-adaptive truncation and step allocation, which significantly reduces the number of reverse diffusion steps at high SNR while maintaining strong performance in low SNR regimes. Simulations on time-evolving standardized channel models demonstrate that the proposed method achieves consistent performance gains over existing diffusion-based channel estimation baselines, while retaining low latency through SNR-adaptive inference.

\end{abstract}

\begin{IEEEkeywords}
MIMO, channel estimation, diffusion models, conditional diffusion, LSTM
\end{IEEEkeywords}

\section{Introduction}

\begin{figure}[t]
    \centering
    \includegraphics[width=0.9\columnwidth]{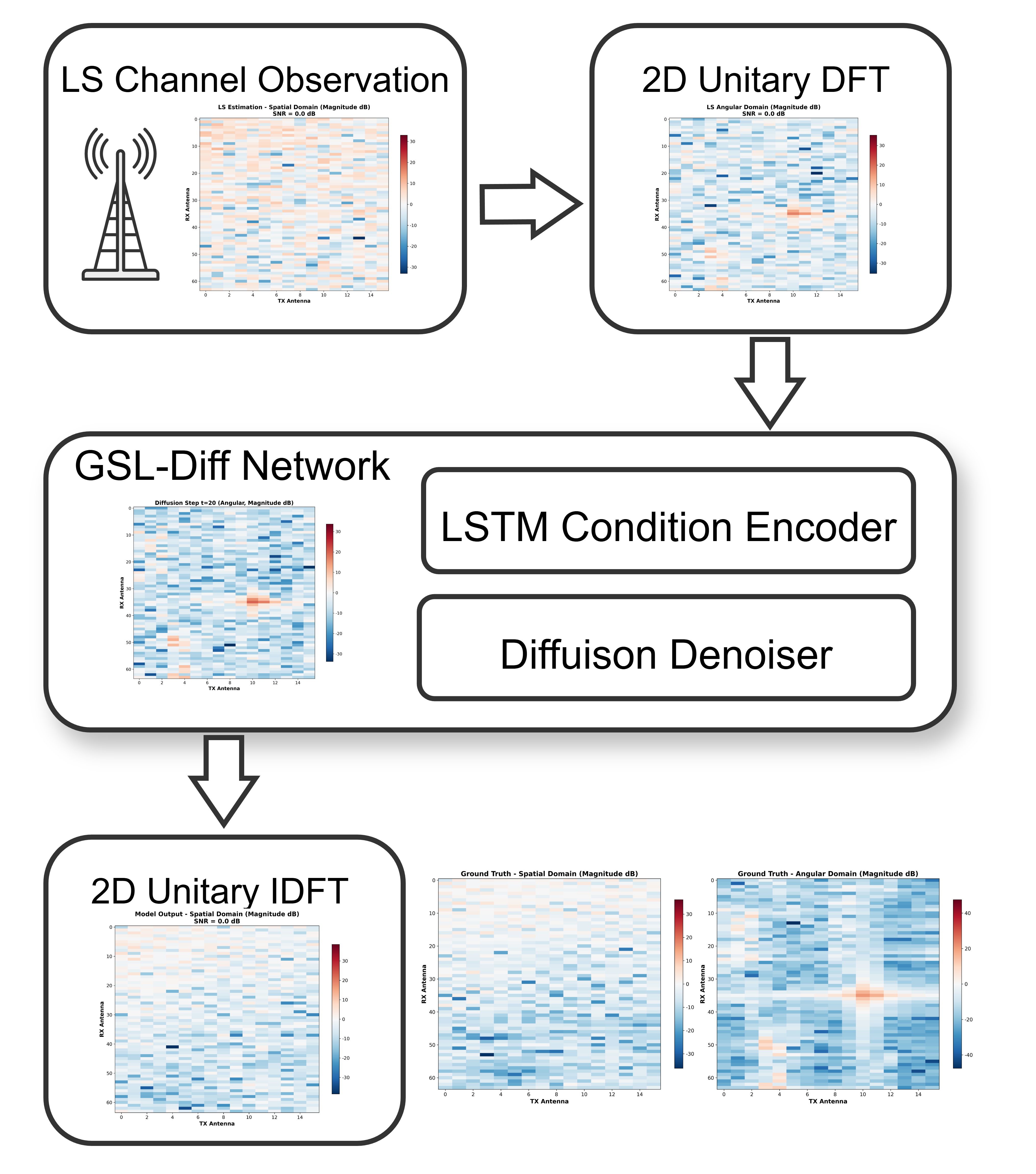}
    \caption{Overall workflow of the proposed channel estimation framework.}
    \label{fig:vtc2026_workflow}
\end{figure}

In 5G/6G massive MIMO systems, the base station relies on accurate channel state information (CSI) to enable key physical layer functionalities. As a result, the accuracy of channel estimation directly determine the achievable system reliability. This challenge becomes more pronounced under emerging 6G scenarios, where enhanced user mobility, higher carrier frequencies (e.g., FR2/FR3), and increasingly large antenna arrays lead to stronger channel time variations and reduced channel coherence. How to achieve \emph{high estimation accuracy, low inference latency, and robustness across a wide SNR range} under limited pilot overhead therefore remains a fundamental open problem in wireless communications.

Recently, data-driven deep learning approaches have been explored to learn a direct regression mapping from noisy pilot observations to channel estimates, showing promising gains especially in low SNR regimes or complex propagation environments. However, purely discriminative models often fail to explicitly capture the underlying channel distribution, and their performance can degrade significantly under distribution shifts. In contrast, diffusion models (DMs), as a class of generative models, are capable of learning complex data distributions through an iterative denoising process. This property has motivated recent works on diffusion-based channel estimation, which combine a learned generative prior with observation consistency to approach MMSE-optimal performance. Existing diffusion-based estimators typically operate in the angular domain on a single channel snapshot and incorporate the SNR as a conditioning variable or positional embedding, achieving improved performance and interpretability compared to conventional methods. Nevertheless, two key limitations remain. First, \emph{temporal correlation is underutilized}: most existing approaches process each time instance independently, failing to exploit the strong inter-frame correlation present in time-varying channels. Second, there exists a fundamental \emph{tradeoff between robustness across SNR and inference latency}: at high SNR, an overly strong generative prior may cause over-smoothing and degrade performance, whereas at low SNR, maintaining accuracy typically requires a larger number of reverse diffusion steps, leading to increased inference time.

To address these challenges, this paper proposes a \emph{time-series-conditioned diffusion framework} for channel estimation, in which the network architecture and inference strategy are jointly optimized to balance performance and complexity. Starting from the LS observation, the estimated channel $\hat{\mathbf{H}}_{\rm LS}$ is first normalized using an SNR-dependent variance scaling to mitigate amplitude variations across different SNRs. A two-dimensional discrete Fourier transform (implemented via FFT2) is then applied along the antenna dimensions to project the channel into the angular domain, where the channel representation becomes more energy-concentrated and structurally sparse, facilitating the learning of generative priors. A conditional diffusion denoiser with a 3D U-Net backbone is employed to predict the clean angular-domain channel $x_0$ during the reverse diffusion process. To explicitly exploit temporal correlation, a dedicated \emph{LSTM-based condition encoder} is introduced to encode a short sequence of LS observations, injecting both global and sequential temporal features into the U-Net via multi-scale cross-attention.

Furthermore, to adaptively balance observation fidelity and generative prior over a wide range of SNR, we introduce a \emph{learnable SNR-gated late-fusion shortcut}. Specifically, the network input $x_{\rm in}$ is re-injected at the decoder output through a sigmoid gate parameterized by the SNR, whose center and scale are learned during training. This mechanism allows the model to preserve fine-grained observation details at high SNR while relying more heavily on the generative prior at low SNR, effectively mitigating over-smoothing effects. Finally, to satisfy low-latency requirements, a deterministic DDIM-style reverse update is adopted together with an \emph{SNR-adaptive truncation and step allocation strategy}. By matching the observation SNR to the effective SNR of the diffusion schedule, the starting timestep $t_{\rm start}$ is determined for each sample, enabling high SNR cases to use fewer reverse steps while automatically allocating more steps to low SNR cases to maintain estimation accuracy.

The main contributions of this work are summarized as follows:
\begin{itemize}
  \item We propose a time-series-conditioned diffusion-based channel estimation framework that performs generative denoising in the angular domain and exploits temporal correlation through an LSTM-encoded sequence of LS observations.
  \item We design a learnable SNR-gated late-fusion shortcut that adaptively balances observation fidelity and generative prior across a wide SNR range, improving robustness in high SNR regimes and alleviating over-smoothing.
  \item Extensive evaluations on standardized time-varying channel datasets demonstrate that the proposed method consistently outperforms existing diffusion-based baselines in terms of NMSE.
\end{itemize}

\section{Proposed Model: GSL-Diffusion}
\label{sec:model}

\begin{figure}[t]
    \centering
    \includegraphics[width=0.95\columnwidth]{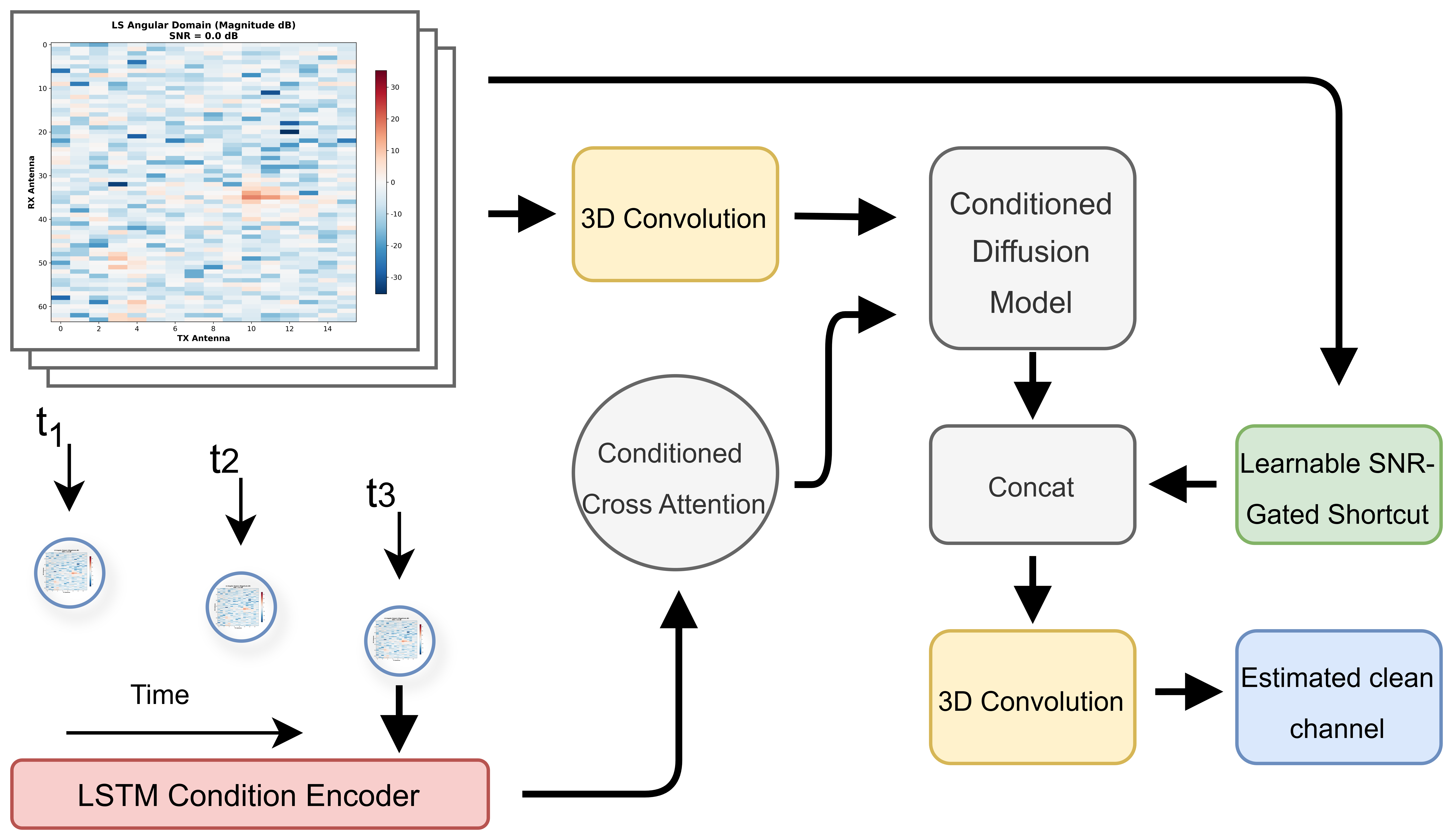}
    \caption{Network architecture of the proposed GSL-Diffusion.}
    \label{fig:network_structure}
\end{figure}

We propose \textbf{GSL-Diffusion} (\,Gated-Shortcut LSTM-Conditioned Diffusion Model\,), a time-series conditioned diffusion estimator that denoises the channel in the angular domain and explicitly balances observation fidelity and learned priors via a learnable SNR-gated late-fusion shortcut.  

\subsection{Preprocessing in the Angular Domain}
Let $\mathbf{H}_{\mathrm{true}}[\tau]\in\mathbb{C}^{N_r\times N_t}$ and $\mathbf{H}_{\mathrm{LS}}[\tau]$ denote the ground-truth and LS channel estimates at time index $\tau\in\{1,\dots,T\}$. We first apply the variance normalization used in prior diffusion estimators:
\begin{equation}
\tilde{\mathbf{H}}_{\mathrm{LS}}[\tau]
= (1+\eta^2)^{-\frac{1}{2}}\,\mathbf{H}_{\mathrm{LS}}[\tau],
\quad
\eta^2=\frac{\kappa}{10^{\mathrm{SNR}_{\mathrm{dB}}/10}},
\label{eq:varnorm}
\end{equation}
where $\kappa$ is the dataset noise-power factor. Then we transform to the sparse angular domain using a 2D unitary FFT over $(N_r,N_t)$:
\begin{equation}
\mathbf{H}^{\mathrm{ang}}[\tau] = \mathcal{F}_2\{\mathbf{H}[\tau]\}.
\end{equation}
Finally, we perform per-sample normalization after the angular transform by dividing both $\mathbf{H}^{\mathrm{ang}}_{\mathrm{true}}$ and $\mathbf{H}^{\mathrm{ang}}_{\mathrm{LS}}$ by a scalar $c=\sqrt{\mathbb{E}[\|\mathbf{H}^{\mathrm{ang}}_{\mathrm{true}}\|_2^2]}$ computed from the current sample. The diffusion model operates on the real-imag stacked tensor
$\mathbf{x}\in\mathbb{R}^{2\times T\times N_r\times N_t}$.

\subsection{Conditional Diffusion Formulation}
We adopt a standard forward diffusion process with $L$ steps:
\begin{equation}
q(\mathbf{x}_t|\mathbf{x}_0)=\mathcal{N}\!\left(\sqrt{\bar{\alpha}_t}\mathbf{x}_0,\ (1-\bar{\alpha}_t)\mathbf{I}\right),
\label{eq:forward}
\end{equation}
where $\bar{\alpha}_t=\prod_{i=1}^t(1-\beta_i)$ and $\beta_i$ denotes the variance added at step $i$.
Equivalently, the noisy sample can be written as
\begin{equation}
\mathbf{x}_t=\sqrt{\bar{\alpha}_t}\mathbf{x}_0+\sqrt{1-\bar{\alpha}_t}\boldsymbol{\epsilon},
\quad
\boldsymbol{\epsilon}\sim\mathcal{N}(\mathbf{0},\mathbf{I}).
\label{eq:xt_def}
\end{equation}

We use the normalized angular-domain LS observation sequence as conditioning, together with the SNR:
\begin{equation}
\mathbf{c} \triangleq \{\mathbf{h}_{\mathrm{LS}},\,\mathrm{SNR}_{\mathrm{dB}}\}.
\label{eq:cond_def}
\end{equation}
Following the ``$x_0$-prediction'' parameterization, the denoiser $f_\theta$ predicts $\mathbf{x}_0$ from $(\mathbf{x}_t,t)$ under condition $\mathbf{c}$:
\begin{equation}
\hat{\mathbf{x}}_0 = f_\theta(\mathbf{x}_t,\ t,\ \mathbf{c}).
\label{eq:x0pred}
\end{equation}
The training objective is the MSE over $\mathbf{x}_0$:
\begin{equation}
\mathcal{L}(\theta)
=\mathbb{E}_{\mathbf{x}_0,t,\boldsymbol{\epsilon}}
\left[
\left\|f_\theta(\mathbf{x}_t,\ t,\ \mathbf{c})-\mathbf{x}_0\right\|_2^2
\right].
\label{eq:loss}
\end{equation}

\subsection{LSTM-Conditioned 3D U-Net Denoiser}
GSL-Diffusion uses a 3D U-Net backbone operating over the $(T,N_r,N_t)$ axes. The condition is extracted from a short sequence of LS observations using a bidirectional LSTM encoder. Concretely, we flatten each LS frame and form a sequence, then apply an input projection and an LSTM with hidden size 512 and 2 layers. The encoder produces:
(i) a global condition vector $\mathbf{c}_g\in\mathbb{R}^{256}$ from the final hidden state, and
(ii) a sequence condition $\mathbf{C}_s\in\mathbb{R}^{T\times 256}$ from the per-timestep outputs.
The diffusion timestep embedding is computed with sinusoidal features and fused with $\mathbf{c}_g$, while $\mathbf{C}_s$ is injected through cross-attention blocks at multiple U-Net resolutions, enabling temporally-aware denoising beyond per-snapshot estimation.

\subsection{Learnable SNR-Gated Late-Fusion Shortcut}
A key design is a \emph{learnable} SNR-gated shortcut that injects the network input into the last decoding stage:
\begin{equation}
g(\mathrm{SNR})
=\sigma\!\left(\frac{\mathrm{SNR}_{\mathrm{dB}}-\mu}{s}\right),
\label{eq:snrgate}
\end{equation}
where $\mu$ (center) and $s$ (scale) are trainable parameters and $\sigma(\cdot)$ is the sigmoid. Let $\mathbf{x}_{\mathrm{in}}=[\mathbf{x}_t;\mathbf{h}_{\mathrm{LS}}]$ denote the channel-wise concatenation (4 input channels). We concatenate $g(\mathrm{SNR})\cdot \mathbf{x}_{\mathrm{in}}$ with the final decoder features before the output convolution. Intuitively, at high SNR the gate tends to favor observation fidelity, while at low SNR it lets the learned diffusion prior dominate.
Intuitively, the proposed gate implements an SNR-aware trust mechanism:
at high SNR, $g(\mathrm{SNR})\!\approx\!1$ so that the estimator preserves observation fidelity,
whereas at low SNR, $g(\mathrm{SNR})\!\approx\!0$ and the denoising relies more heavily on the learned generative prior.

\subsection{SNR-Adaptive Truncated DDIM Inference}
To reduce latency, we adopt deterministic DDIM-style reverse updates and \emph{truncate} the reverse trajectory according to the observation SNR. We match the observation SNR to the diffusion SNR schedule
$\mathrm{SNR}_{\mathrm{DM}}(t)=\bar{\alpha}_t/(1-\bar{\alpha}_t)$ via:
\begin{equation}
t_{\mathrm{start}}=\arg\min_t \left|\mathrm{SNR}_{\mathrm{dB}}-\mathrm{SNR}_{\mathrm{DM,dB}}(t)\right|.
\label{eq:tstart}
\end{equation}
We then allocate the number of DDIM steps by
\begin{equation}
K=\mathrm{clip}\!\left(\left\lfloor \frac{t_{\mathrm{start}}}{L-1}K_{\max}\right\rceil,\ K_{\min},\ K_{\max}\right),
\label{eq:ksteps}
\end{equation}
and generate an evenly spaced sequence from $t_{\mathrm{start}}$ to 0. We initialize with $\mathbf{x}_{t_{\mathrm{start}}}=\mathbf{h}_{\mathrm{LS}}$ (normalized angular LS), and iterate deterministically:
\begin{align}
\hat{\boldsymbol{\epsilon}}_t
&=\frac{\mathbf{x}_t-\sqrt{\bar{\alpha}_t}\hat{\mathbf{x}}_0}{\sqrt{1-\bar{\alpha}_t}+\epsilon},\\
\mathbf{x}_{t'}
&=\sqrt{\bar{\alpha}_{t'}}\hat{\mathbf{x}}_0+\sqrt{1-\bar{\alpha}_{t'}}\hat{\boldsymbol{\epsilon}}_t,
\label{eq:ddim}
\end{align}
where $t'$ is the next (smaller) timestep and $\epsilon$ is a small constant for numerical stability. After sampling, we undo sample normalization and apply inverse FFT to obtain the spatial-domain estimate.

\begin{algorithm}[t]
\caption{SNR-adaptive truncated DDIM inference}
\label{alg:adaptive_ddim}
\begin{algorithmic}[1]
\STATE \textbf{Input:} normalized angular LS $\mathbf{h}_{LS}$, SNR$_{dB}$, schedule $\{\bar\alpha_t\}_{t=0}^{L-1}$, bounds $[K_{\min},K_{\max}]$
\STATE Compute diffusion SNR: $\mathrm{SNR}_{DM}(t)=\bar\alpha_t/(1-\bar\alpha_t)$
\STATE $t_{\mathrm{start}} \leftarrow \arg\min_t |\mathrm{SNR}_{dB}-10\log_{10}(\mathrm{SNR}_{DM}(t))|$
\STATE $K \leftarrow \mathrm{clip}\big(\mathrm{round}(\frac{t_{\mathrm{start}}}{L-1}K_{\max}), K_{\min}, K_{\max}\big)$
\STATE Build an evenly spaced timestep sequence $\{t_k\}_{k=0}^{K}$ from $t_{\mathrm{start}}$ to $0$
\STATE Initialize $\mathbf{x}_{t_{\mathrm{start}}}\leftarrow \mathbf{h}_{LS}$
\FOR{$k=0,\dots,K-1$}
  \STATE $\hat{\mathbf{x}}_0 \leftarrow f_\theta(\mathbf{x}_{t_k}, t_k, \mathbf{h}_{LS}, \mathrm{SNR}_{dB})$
  \STATE $\hat{\boldsymbol{\epsilon}} \leftarrow (\mathbf{x}_{t_k}-\sqrt{\bar\alpha_{t_k}}\hat{\mathbf{x}}_0)/\sqrt{1-\bar\alpha_{t_k}}$
  \STATE $\mathbf{x}_{t_{k+1}} \leftarrow \sqrt{\bar\alpha_{t_{k+1}}}\hat{\mathbf{x}}_0+\sqrt{1-\bar\alpha_{t_{k+1}}}\hat{\boldsymbol{\epsilon}}$
\ENDFOR
\STATE \textbf{Output:} $\hat{\mathbf{x}}_0$ (undo normalization + IFFT to spatial domain)
\end{algorithmic}
\end{algorithm}

\subsection{Implementation Details}
We train GSL-Diffusion with $L{=}100$ diffusion steps using a linear $\beta$ schedule ($\beta_1{=}10^{-3}$ to $\beta_L{=}0.08$) and $x_0$-prediction in the angular domain. 
The condition encoder is a 2-layer bi-LSTM (hidden size 512) followed by a projection to 256-dim global and sequential conditions, injected into a 3D U-Net via multi-scale cross-attention. 
We optimize with AdamW (lr $10^{-4}$, weight decay $10^{-4}$) for 200 epochs, batch size 32, gradient clipping 1.0, and a warmup + cosine learning-rate schedule. 
During training, we apply complex phase-rotation and spatial flip augmentations to improve invariance.

% \begin{table}[t]
% \centering
% \caption{Key implementation settings of GSL-Diffusion.}
% \label{tab:impl}
% %\scriptsize
% \setlength{\tabcolsep}{4pt}
% \begin{tabular}{l l}
% \hline
% \textbf{Component} & \textbf{Setting} \\
% \hline
% Diffusion steps & $L=100$, linear $\beta$: $10^{-3}\rightarrow 0.08$ \\
% Parameterization & $x_0$-prediction, DDIM sampling \\
% U-Net backbone & 3D U-Net, base ch.=64, mult=(1,2,4), 2 res-blocks \\
% Condition encoder & bi-LSTM, 2 layers, hidden=512, cond dim=256 \\
% Condition injection & multi-scale cross-attention (encoder+decoder) \\
% Optimizer & AdamW, lr=$10^{-4}$, wd=$10^{-4}$ \\
% Training & 200 epochs, batch=32, grad clip=1.0, warmup+cosine \\
% Augmentation & random complex phase rotation + spatial flips \\
% \hline
% \end{tabular}
% \end{table}

\section{Dataset and Evaluation Setup}
\label{sec:dataset_eval}

\begin{figure*}[t]
    \centering
    \includegraphics[width=\textwidth]{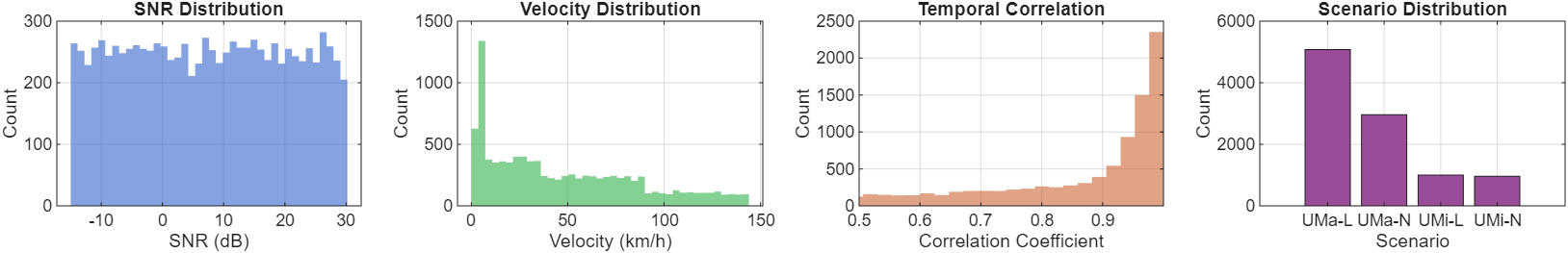}
    \caption{Distribution of the QuaDRiGa dataset: (a) SNR (b) Velocity 
(c) Temporal correlation coefficient $\bar{\rho}$ between adjacent snapshots (d) Scenario distribution.}
    \label{fig:quadriga_dataset_distributions}
\end{figure*}

\subsection{QuaDRiGa-Based Time-Varying MIMO Dataset}
We generate a standardized time-varying MIMO dataset using QuaDRiGa under 3GPP TR~38.901 urban scenarios. Each sample provides the ground-truth channel sequence, a physically consistent pilot observation, the LS initialization, and meta-data (scenario/velocity/Doppler) for grouped analysis.

\subsubsection{System and array configuration}
We consider a narrowband $N_{\rm RX}\!\times\!N_{\rm TX}=64\!\times\!16$ MIMO link at $f_c=6$~GHz.
Both BS and UE employ omni, single-polarized ULAs with inter-element spacing $\lambda/2$.
QuaDRiGa is configured with 100~MHz bandwidth and generates frequency responses on $N_{\rm sc}=N_{\rm TX}$ subcarriers; we average over subcarriers to obtain an equivalent narrowband channel matrix, focusing on spatial/angle-domain structure.

\subsubsection{3GPP~38.901 scenarios (UMa/UMi, LOS/NLOS)}
We mix four standardized 3GPP~38.901 scenario types:
UMa-LOS, UMa-NLOS, UMi-LOS, and UMi-NLOS (default ratio 0.5/0.3/0.1/0.1).  
UMa (Urban Macro) represents macro BS deployment (BS above rooftops) with $h_{\rm BS}=25$~m, while
UMi (Urban Micro) represents street-level micro BS deployment with $h_{\rm BS}=10$~m (UE height 1.5~m for all cases).
The UE horizontal distance is uniformly sampled within scenario-dependent ranges: UMa: $d_{\rm 2D}\!\in\![35,500]$~m; UMi: $d_{\rm 2D}\!\in\![10,200]$~m, and azimuth angles are uniformly randomized to cover diverse AoA/AoD conditions.
LOS indicates the presence of a direct path between BS and UE, typically yielding a stronger specular component, whereas
NLOS indicates that the direct path is blocked and propagation is dominated by reflection/diffraction/scattering, resulting in richer multipath and higher attenuation.
In our dataset, LOS/NLOS are \emph{explicitly selected} via scenario types to control the composition (rather than relying on a distance-dependent LOS probability).

\subsubsection{Mobility and temporal correlation}
Each sample contains $T=14$ snapshots with interval $\Delta t=1$~ms. \textit{(Stored as a time sequence in the dataset tensors.)}
UE speed $v$ is sampled from four mobility ranges
$[0.5,2]$, $[2,10]$, $[10,25]$, $[25,40]$~m/s with weights (0.2/0.3/0.35/0.15), covering pedestrian to high-mobility regimes.
We compute Doppler as $f_d=v f_c/c$ and enable QuaDRiGa drifting mode to generate smoothly time-evolving channels across snapshots.

To quantify temporal dynamics, we measure the (magnitude) normalized adjacent-snapshot correlation
\begin{equation}
\bar{\rho}
=\frac{1}{T-1}\sum_{t=1}^{T-1}
\frac{\left|\left\langle \mathbf{H}_t, \mathbf{H}_{t+1}\right\rangle_F\right|}
{\left\|\mathbf{H}_t\right\|_F \left\|\mathbf{H}_{t+1}\right\|_F},
\label{eq:temp_corr}
\end{equation}
where $\langle \mathbf{A},\mathbf{B}\rangle_F=\mathrm{tr}(\mathbf{A}^H\mathbf{B})$.
As shown in Fig.~\ref{fig:quadriga_dataset_distributions}(c), $\bar{\rho}$ spans roughly $[0.5,1]$ but the majority of samples are highly correlated and concentrate near $1$, indicating predominantly slowly time-varying channels with a non-negligible faster-varying tail. This provides learnable temporal structure for the LSTM-conditioned diffusion estimator.

\subsubsection{Pilot observation and LS initialization}
Rather than injecting noise directly on channel coefficients, we explicitly simulate pilot transmission.
Let $\mathbf{H}_t\!\in\!\mathbb{C}^{N_{\rm RX}\times N_{\rm TX}}$ denote the channel at snapshot $t$ and
$\mathbf{P}\!\in\!\mathbb{C}^{N_{\rm TX}\times N_{\rm TX}}$ be a unitary DFT pilot with $\mathbf{P}\mathbf{P}^H=\mathbf{I}$.
The received pilot observation and LS estimate are
\begin{equation}
\mathbf{Y}_t=\mathbf{H}_t\mathbf{P}+\mathbf{N}_t,\qquad
\widehat{\mathbf{H}}^{\rm LS}_t=\mathbf{Y}_t\mathbf{P}^H,
\label{eq:pilot_ls}
\end{equation}
where $\mathbf{N}_t$ has i.i.d.\ entries $\sim\mathcal{CN}(0,\sigma^2)$.
Because $\mathbf{P}$ is unitary, $\mathbf{N}_t\mathbf{P}^H$ preserves the same noise statistics, matching classical LS modeling while remaining physically consistent.

\subsubsection{Dataset size and split}
We use 50{,}000 training samples and 10{,}000 test samples. Channels are stored in real/imag form for all $T$ snapshots, together with $\text{SNR}_{\rm dB}$, velocity, Doppler, and scenario labels.

\subsection{Evaluation Protocol}
\subsubsection{Metric}
We report NMSE (in dB):
\begin{equation}
\mathrm{NMSE_{dB}}
=10\log_{10}\!\left(\mathbb{E}\left[\frac{\|\mathbf{H}-\widehat{\mathbf{H}}\|_F^2}{\|\mathbf{H}\|_F^2}\right]\right).
\label{eq:nmse}
\end{equation}
Results are shown versus SNR and can be further grouped by mobility (velocity/Doppler) and scenario.

\subsubsection{Baselines and timing}
We compare with LS and LMMSE baselines, as well as diffusion-model ablations (different conditioners/shortcuts) under the same sampling strategy.
Inference latency is measured on GPU with a warm-up stage followed by repeated timed runs; for diffusion estimators we adopt deterministic DDIM-style updates and SNR-adaptive step truncation to reduce sampling iterations at high SNR.

\section{Results}
\label{sec:results}

\begin{figure}[t]
  \centering
  \includegraphics[width=0.45\textwidth]{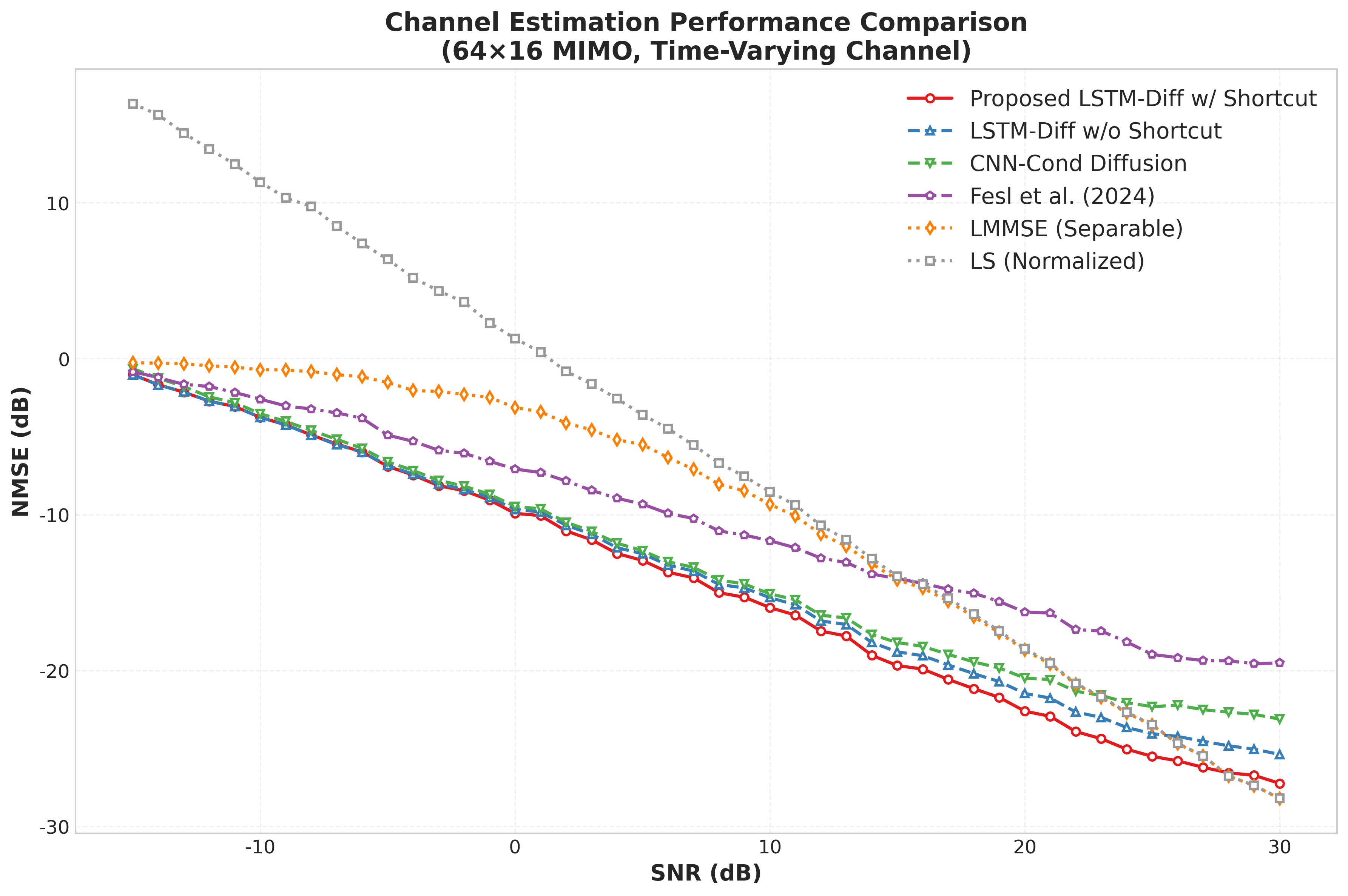}
  \caption{NMSE vs.\ SNR for time-varying $64\times 16$ MIMO channel estimation.}
  \label{fig:nmse}
\end{figure}

\subsection{NMSE vs.\ SNR}
Fig.~\ref{fig:nmse} shows that the proposed GSL-Diffusion achieves the best NMSE almost across the full SNR range from $-15$ to $30$\,dB. Under the \emph{same} diffusion denoiser backbone, replacing the LSTM encoder with a CNN encoder leads to inferior high SNR behavior and becomes worse than separable LMMSE beyond about $23$\,dB. Removing the SNR-gated late-fusion shortcut from the LSTM-conditioned model also degrades the high SNR regime, falling behind LMMSE beyond about $26$\,dB. A diffusion model that uses the SNR-gated shortcut but lacks the full LSTM temporal conditioning becomes worse than LMMSE beyond about $28$\,dB. Overall, the full GSL-Diffusion is consistently optimal, indicating that \emph{both} temporal conditioning and the learnable SNR-gated shortcut are necessary to avoid over-smoothing at high SNR while retaining strong generative denoising at low SNR.

\begin{figure}[t]
  \centering
  \includegraphics[width=0.45\textwidth]{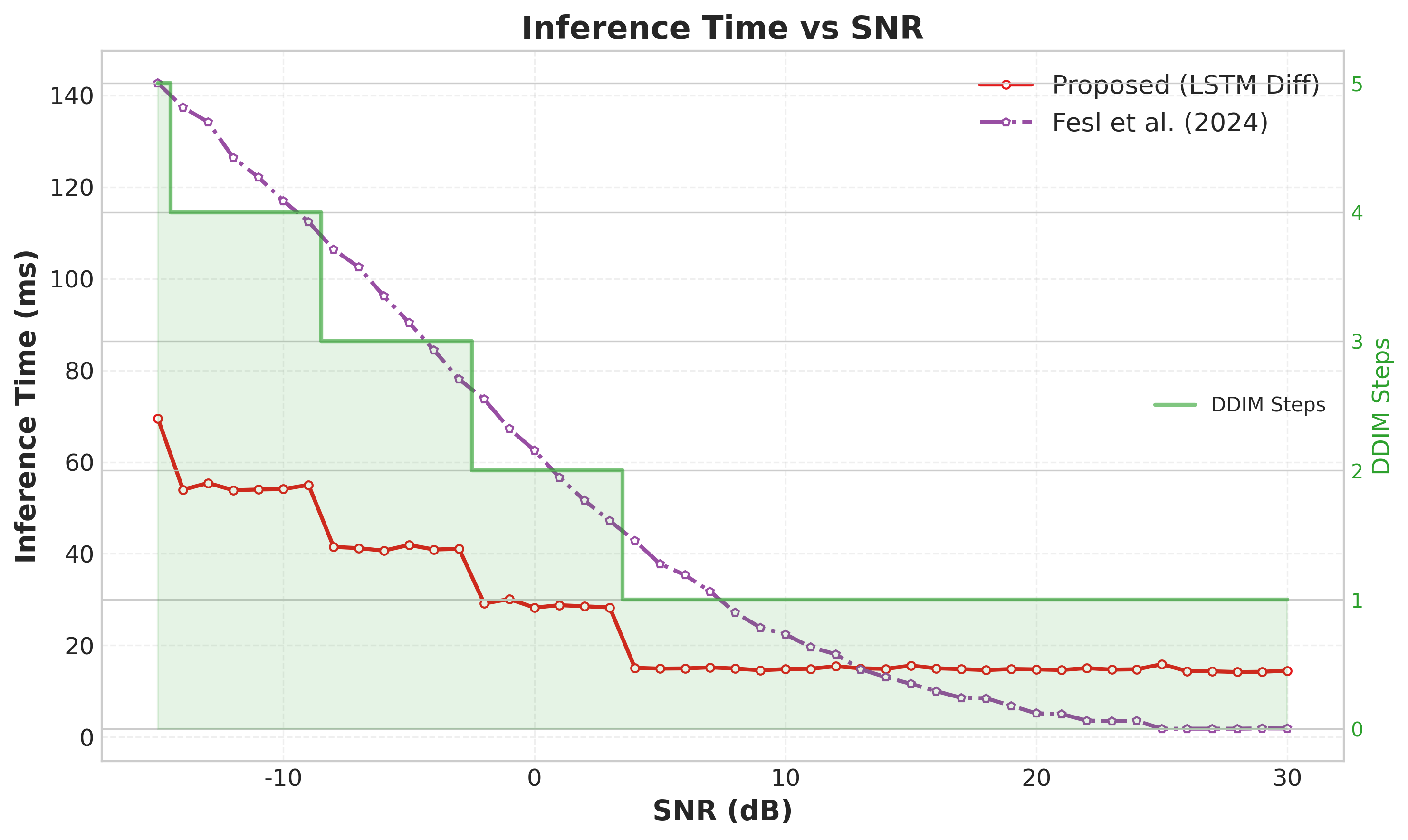}
  \caption{Inference time vs.\ SNR. The secondary axis shows the SNR-adaptive DDIM step count.}
  \label{fig:time}
\end{figure}

\subsection{Inference Time vs.\ SNR}
Fig.~\ref{fig:time} reports the end-to-end inference latency measured on a workstation equipped with a AMD 9950X CPU and an NVIDIA RTX5090 GPU. 
For each SNR point in $[-15,30]$\,dB (1\,dB resolution), we adopt an online-style batch size of 1 and measure runtime after a warm-up stage, averaging 50 repeated timed runs with GPU synchronization to reduce measurement noise.

Although our proposed GSL-Diffusion backbone is substantially larger than the lightweight diffusion baseline of Fesl \emph{et al.}, resulting in a higher per-step runtime (13.2\,ms/step for ours vs.\ 1.7\,ms/step for Fesl on the same GPU), the overall latency is dominated by the \emph{number of reverse updates} rather than the per-step cost.
Thanks to the SNR-adaptive truncated DDIM strategy, the step count is automatically reduced at high SNR (often reaching the minimum of 1 step), while allocating more steps only when the observation is heavily corrupted.
As illustrated by the secondary axis in Fig.~\ref{fig:time}, this adaptive allocation yields low delay in favorable conditions and maintains bounded, practically acceptable latency under strong interference/low SNR regimes by capping the maximum number of steps ($K_{\max}$).
Overall, the proposed method achieves a better accuracy--latency tradeoff: it tolerates a heavier denoiser per update but compensates with fewer required updates to reach the target NMSE.

\section{Conclusion}
This paper presented GSL-Diffusion, an SNR-adaptive, time-series-conditioned diffusion model for MIMO channel estimation that performs denoising in the angular domain. By encoding short LS observation sequences with an LSTM and injecting them into a 3D U-Net via multi-scale conditioning, the proposed estimator explicitly exploits temporal correlation beyond per-snapshot processing. A learnable SNR-gated late-fusion shortcut was introduced to adaptively trade off observation fidelity and generative priors, mitigating over-smoothing in high SNR regimes while preserving strong denoising at low SNR. To satisfy latency constraints, we further employed deterministic DDIM updates with SNR-adaptive truncation and step allocation, substantially reducing reverse steps at high SNR. Experiments on standardized time-varying channel datasets demonstrated consistent NMSE gains over LS/LMMSE and diffusion-based baselines, alongside an improved accuracy--latency tradeoff.

\end{document}